\documentclass[runningheads]{llncs}
\usepackage[T1]{fontenc}
\usepackage{graphicx,verbatim}
\usepackage{mathrsfs}
\usepackage{multirow}
\usepackage{hyperref}
\usepackage{amsmath} 
\begin{document}
\title{DDO-IN: Dual Domains Optimization for Implicit Neural Network to Eliminate Motion Artifact in Magnetic Resonance Imaging}
%
\author{Zhongyu Mai\inst{1} \and
Zewei Zhan\inst{1} \and
Hanyu Guo\inst{1}  \and
Yulang Huang\inst{1}    \and
Weifeng Su\inst{1,2}}
%
%
\institute{Department of Computer Science, BNU-HKBU United International College \and
Guangdong Provincial Key Laboratory of IRADS }

\maketitle              
\begin{abstract}
Magnetic resonance imaging (MRI) motion artifacts can seriously affect clinical diagnostics, making it challenging to interpret images accurately. Existing methods for eliminating motion artifacts struggle to retain fine structural details and simultaneously lack the necessary vividness and sharpness. In this study, we present a novel dual-domain optimization (DDO) approach that integrates information from the pixel and frequency domains guiding the recovery of clean magnetic resonance images through implicit neural representations(INRs). Specifically, our approach leverages the low-frequency components in the k-space as a reference to capture accurate tissue textures, while high-frequency and pixel information contribute to recover details. Furthermore, we design complementary masks and dynamic loss weighting transitioning from global to local attention that effectively suppress artifacts while retaining useful details for reconstruction. Experimental results on the NYU fastMRI dataset demonstrate that our method outperforms existing approaches in multiple evaluation metrics. Our code is available at https://anonymous.4open.science/r/DDO-IN-A73B.

\end{abstract}
\section{Introduction}
Magnetic Resonance Imaging (MRI) constitutes a noninvasive medical imaging modality rooted in the principles of nuclear magnetic resonance. This technique is notably effective in differentiating soft tissues, which renders it an essential instrument in the clinical diagnosis of neurological disorders, tumor identification, and functional imaging. The procedure necessitates extended signal acquisition periods, which may span from several minutes to numerous tens of minutes. During this interval, patient motion, whether voluntary or involuntary (including acts such as respiration, deglutition, or tremor), can provoke temporal and spatial inconsistencies in signal phase or frequency, culminating in image blurring, ghosting, or stripe distortions. The mitigation of these artifacts not only precludes the obscuration of lesions or the occurrence of misdiagnoses, thereby enhancing diagnostic precision, but also diminishes the requirement for repeated scans. This, in turn, contributes to a reduction in healthcare expenditures and augments patient comfort\cite{27b9201ad4e24e51a7cc25724c536a80,029855cdf7ac4cfa83c8c42d60898650}.

In the domain of MRI, the elimination of artifacts has emerged as an essential endeavor to enhance both the quality of the images produced and the reliability of subsequent diagnoses. To confront this challenge, researchers have developed a spectrum of strategies. Techniques for prospective prevention of motion-induced artifacts aim to preemptively mitigate these disruptions during the MRI procedure by implementing real-time tracking and adjustments. Such methods are designed to minimize the influence of artifacts ahead of time by dynamically modifying imaging sequences and parameters in response to the patient’s movements. Nevertheless, these approaches frequently necessitate additional apparatus, which can result in prolonged scanning durations. This both elevates the operational expenses and adds to patient discomfort, representing significant considerations in clinical settings. \cite{Schulz2012AnEO,ZAITSEV20061038,https://doi.org/10.1002/mrm.26274}. Generative Adversarial Networks (GANs)-based methods have also emerged in artifact removal research, including approaches such as Pix2pix and CycleGAN, which repair motion artifacts through supervised or unsupervised learning \cite{ARMANIOUS2020101684,Li_SelfsupervisedDenoising_MICCAI2024}. These methods utilize a generator and discriminator to restore normal structures in the affected regions. However, due to the diverse nature of motion artifact distribution and the varying types of motion in clinical settings, GANs frequently struggle to handle these uncertainties, leading to inconsistent restoration outcomes. Finally, diffusion-based models, such as DR2 and GDP, have gained considerable attention in recent years \cite{Chung2021SimultaneousSA,Si2023FreeUFL}. These models simulate the image generation process to restore motion-affected images. While these methods perform well in image generation, they still confront challenges in motion artifact removal, particularly in terms of addressing phase disturbances in the k-space. Diffusion models focus mainly on information in the pixel domain, neglecting the high frequency perturbations typical of motion artifacts in the k space, leading to inaccurate or distorted artifact removal, especially during significant patient movement \cite{Song2020DenoisingDI,Ho2022ClassifierFreeDG}. Recent developments in implicit neural representations (INRs) have been successfully applied to self-supervised MRI reconstruction, offering an innovative approach to improving motion artifact removal, although there is still potential for further enhancement \cite{hemidi2024immocoselfsupervisedmrimotion,10356136,Huang2022NeuralIK}. 

This study introduces a novel DDO-IN technique that leverages implicit neural networks to proficiently eliminate motion-induced artifacts, simultaneously maintaining the integrity of structural data and preserving intricate textural details. The principal contributions of this work can be summarized as follows: 1) We establish a dual-domain optimization pipeline tailored for implicit neural networks (DDO-IN) and design parameters to equilibrate the weight within frequency and pixel domains. 2) We propose an ingenious loss function that incorporates a complementary mask operation alongside dynamic weighting to enhance the model's performance. 3) The efficacy of the proposed methods is validated through experiments on the NYU fastMRI dataset, demonstrating significant metric improvements.

\begin{figure}[h!]
\centering
\includegraphics[width=\textwidth]{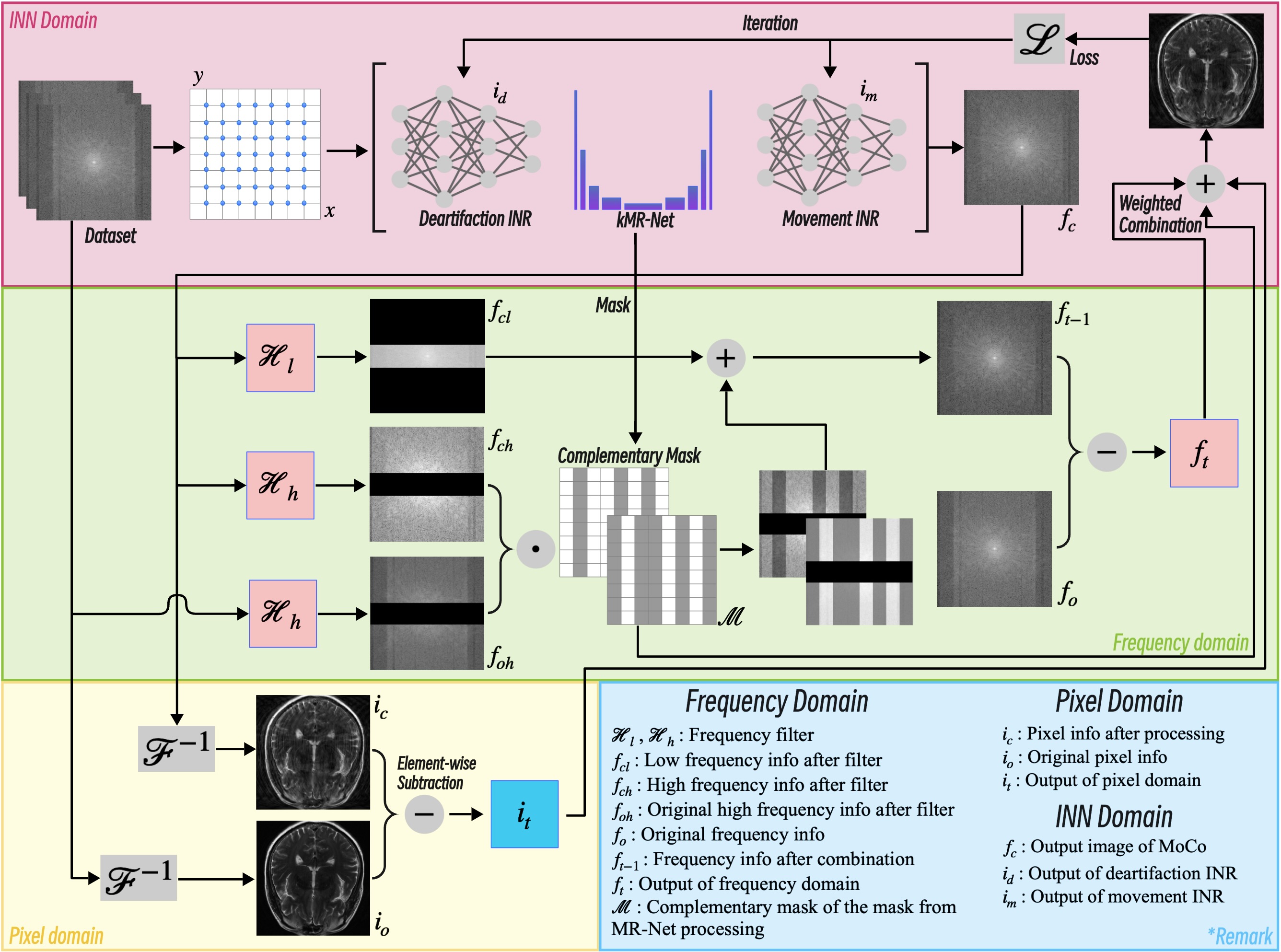}
\caption{The framework of DDO-IN. The red area corresponds to the INN module, responsible for generating the deartifacted image and motion transformation map. These two outputs are subsequently fused via the kMR-Net. The green area illustrates frequency domain processing, which combines low-frequency and high-frequency information. The yellow area indicates pixel domain processing, capturing pixel-level details. After merging the results from both domains, the clean image is generated by Deartifact INR through iterative optimization of the loss function.}
\label{fig:motioncorrectionresults}
\end{figure}

\section{Methodology}
This is our overview, as illustrated in Fig.~\ref{fig:motioncorrectionresults}. The process is as follows: 1) K-space data enter an INN framework, yielding an artifact-corrupted image. 2) A k-space Mask Recognition Net (kMR-Net) generated binary mask aids the INR and serves as the complementary mask later. 3) The loss in the frequency domain combines the generated and original high-frequency components with the generated low-frequency components. The loss in the pixel domain compares the inverse Fourier transforms of the generated and original images. 4) Dynamic weighting adjusts model attention between regions during training.

\subsection{Implicit Neural Network}
The network is comprised of two principal modules: the Deartifact INR and the Movement INR. The primary aim of the Deartifact INR is to effectively eliminate motion artifacts from MRI images while maintaining their structural features. The network utilizes implicit neural representations (INRs) employing hash grid encoding to encapsulate the structural characteristics linked with artifacts, culminating in the generation of an artifact-free image, denoted as \(i_d\). The central function of the Movement INR is to estimate image deformations induced by motion. This method involves learning motion information from motion artifact images using n-dimensional hash grid encoding and a learnable multi-layer perceptron (MLP). This procedure results in the creation of several motion transformation maps, indicated as image \(i_m\). Ultimately, a Fourier transform (FFT) is executed on this image, after which it is integrated with the mask for further processing. This procedure yields a simulated motion-corrupted frequency domain image, referred to as \(f_{c}\). The subsequent process is described in detail as follows:
\begin{equation}
    f_{c}=\mathscr{F}(i_d) \cdot (1 - \mathscr{M}) + \mathscr{F}(i_m) \cdot \mathscr{M} 
\end{equation}

where \(\mathscr{M}\) represents masks, further elucidated subsequently.

\subsection{Dual Domain Optimization Pipeline}
\textbf{Mask Acquisition} In the initial phase of our pipeline, we introduce a k-space Motion Recognition Network (kMR-Net) designed to identify lines corrupted by motion in the raw k-space data. The kMR-Net architecture is derived from the Unet model\cite{Ronneberger2015UNetCN}. Utilizing the trained kMR-Net model, we produce masks for motion artifacts. Throughout the training procedure, we establish a threshold condition whereby a column is deemed to contain motion artifacts if the proportion of such artifacts exceeds 30\%, leading to the output of a corresponding mask.

\textbf{Dual Domain Optimization} The initial focus is on the removal of motion artifacts within the frequency domain. The process commences with the extraction of low-frequency components from the simulated motion-corrupted image described as:
\begin{equation}
    f_{cl} = \mathscr{H}_l(f_c)
\end{equation}
This process contributes to the facilitation of subsequent image reorganization. Following this, the complementary mask, produced by the kMR-Net model, is amalgamated with the high-frequency region of both the simulated motion-degraded image and the original corrupted image. This integration is further combined with the original low-frequency data to achieve the finalized reorganized image. 
\begin{equation}
    f_{ic} =f_{cl} + f_{ch} \cdot  \mathscr{M}  + f_{oh} \cdot (1 -  \mathscr{M} )
\end{equation}
Subsequently, the inverse Fourier transform (IFFT) is employed to transform the image from the frequency domain to the pixel domain, resulting in a motion-corrected and artifact-free image:
\begin{equation}
    i_{\text{o}} = \mathscr{F}^{-1} \left( f_{\text{o}} \right)
\end{equation}
\begin{equation}
    i_c = \mathscr{F}^{-1} \left( f_c \right)
\end{equation}
Ultimately, the images generated, as referenced in \(i_{\text{o}}\) and \(i_{\text{dm}}\), are introduced into the loss function module for additional optimization. 

\subsection{Dynamic Loss Function}
A dynamical weighting mechanism is introduced during the training phase of the model. This weighting parameter (\(\omega\)) decreases linearly from 0.5 to 0 and is integral to two principal aspects: the computation of loss in the frequency domain mask and the equilibrium between losses in the frequency and pixel domains.
\begin{equation}
    \omega = -\frac{t}{2T} + 0.5
\end{equation}
Where t represents the current training epoch and T represents the total number of training epochs. Mean Squared Error (MSE) is applied to calculate losses in both frequency and pixel domains. 
\begin{equation}
    {Loss}_{\text{freq}} = \frac{1}{N} \sum_{i=1}^{N} \left( \hat{F}(x_i) - F(x_i) \right)^2
\end{equation}
\begin{equation}
    {Loss}_{\text{pixel}} = \frac{1}{M} \sum_{i=1}^{M} \left( \hat{I}(y_i) - I(y_i) \right)^2
\end{equation}
Where \( \text{Loss}_{freq} \) and \( \text{Loss}_{pixel} \) represent the frequency domain and pixel domain losses, respectively. \( N \) denotes the number of frequency components, while \( M \) represents the number of pixels. \( \hat{F}(x_i) \) and \( F(x_i) \) refer to the predicted and actual frequency domain values at position \( x_i \), respectively. Similarly, \( \hat{I}(y_i) \) and \( I(y_i) \) are the predicted and actual pixel intensities at position \( y_i \), respectively. 

Subsequently, the frequency domain loss is multiplied by the mask and the weight (\(\omega\)) to dynamically adjust the model's attention towards regions affected by artifacts and those that are artifact-free. During the initial training stages, \(\omega\) is configured to 0.5, ensuring that the model learns features from both regions equally. As training advances, \(\omega\) is progressively reduced to 0, leading the model to concentrate more on detailed restoration in artifact-free regions. This dynamic adjustment mechanism facilitates the model in initially acquiring overall image features, followed by the refinement of fine details. The corresponding formula is applied:
\begin{equation}
    {Loss}_{w\_freq} = \omega \cdot ({Loss}_{freq} \cdot \mathscr{M}) + (1 - \omega) \cdot ({Loss}_{freq} \cdot(1 - \mathscr{M}))
\end{equation}
Eventually, the collective total loss is derived by aggregating the weighted frequency domain loss and the pixel domain loss, which subsequently directs the parameter updates of the model. 
\begin{equation}
    {Loss} = \omega \cdot {Loss}_{w\_freq} + (1 - \omega) \cdot {Loss}_{pixel}
\end{equation}
This dynamic weight adjustment mechanism serves to enhance the model's efficacy in artifact removal and concurrently augments the quality of overall image reconstruction.

\section{Experimental result}
\subsection{Datasets \& Setup}
In this study, we utilize the NYU fastMRI dataset. This dataset comprises 6,970 fully sampled brain MRIs acquired on 1.5T and 3T magnets, including axial T1-weighted, T2-weighted, and FLAIR images\cite{Zbontar2018fastMRIAO}. The data is stored in HDF5 format, with each file representing a single MRI acquisition. For analysis, we randomly divide the dataset into training (200 sequences), validation (50 sequences), and test (50 sequences) sets. Each MRI sequence was standardized to a size of 320×320 pixels. We create paired datasets with and without artifacts. Specifically, we introduce variations including random rotations and translational movements of up to ±10 mm, to mimic both light artifacts (featuring 6-10 instances of motion) and more pronounced artifacts (exhibiting 16-20 instances of motion). The training of the model entailed 250 epochs conducted with the Adam optimization algorithm, where the initial learning rate was set at 5e-4.

\begin{table}[ht]
\caption{Quantitative evaluation of experiment I: motion correction across all methods and motion scenarios. Results are reported as mean ± standard deviation over the test set. All metrics, except PSNR (dB), are reported as percentages. Arrows indicate the direction of improvement.}
\label{tab1}
\resizebox{\textwidth}{!}{
  \begin{tabular}{|l|l|l|l|l|l|}\hline
    Scenario & Method & VIF & HaarPSI & PSNR & SSIM \\
  \hline
  \multirow{5}{*}{Light} & Motion Corrupted & 38.22$\pm$9.07 & 67.94$\pm$8.71 & 26.56$\pm$3.66 & 81.66$\pm$7.43 \\
  & UNet & 43.53$\pm$7.39 & 73.01$\pm$8.11 & 27.87$\pm$3.93 & 88.85$\pm$3.48 \\
  & AF & 63.16$\pm$8.44 & 86.27$\pm$6.44 & 32.93$\pm$4.02 & 93.83$\pm$2.91 \\
  & IM-MoCo & 75.98$\pm$9.38 & 94.54$\pm$5.74 & 37.77$\pm$5.03 & 97.26$\pm$2.20 \\
  & \textbf{DDO-IN} & \textbf{77.34$\pm$8.79} & \textbf{95.76$\pm$4.99} & \textbf{38.68$\pm$4.88} & \textbf{97.42$\pm$2.17} \\
  \hline
  \multirow{5}{*}{Heavy} & Motion Corrupted & 24.35$\pm$8.10 & 53.98$\pm$8.18 & 22.99$\pm$2.73 & 66.01$\pm$11.19 \\
  & UNet & 30.71$\pm$7.11 & 60.20$\pm$7.88 & 25.45$\pm$3.71 & 80.71$\pm$4.88 \\
  & AF & 45.75$\pm$6.95 & 74.13$\pm$6.58 & 28.09$\pm$2.69 & 86.18$\pm$3.74 \\
  & IM-MoCo & 58.57$\pm$10.70 & 87.34$\pm$8.44 & 32.89$\pm$4.28 & 93.01$\pm$3.33 \\
  & \textbf{DDO-IN} & \textbf{60.15$\pm$10.86} & \textbf{89.30$\pm$7.87} & \textbf{33.33$\pm$4.25} & \textbf{93.28$\pm$3.26} \\
  \hline
  \end{tabular}
}
\end{table}

\begin{figure}[h!]
	\centering
	\includegraphics[width=\textwidth]{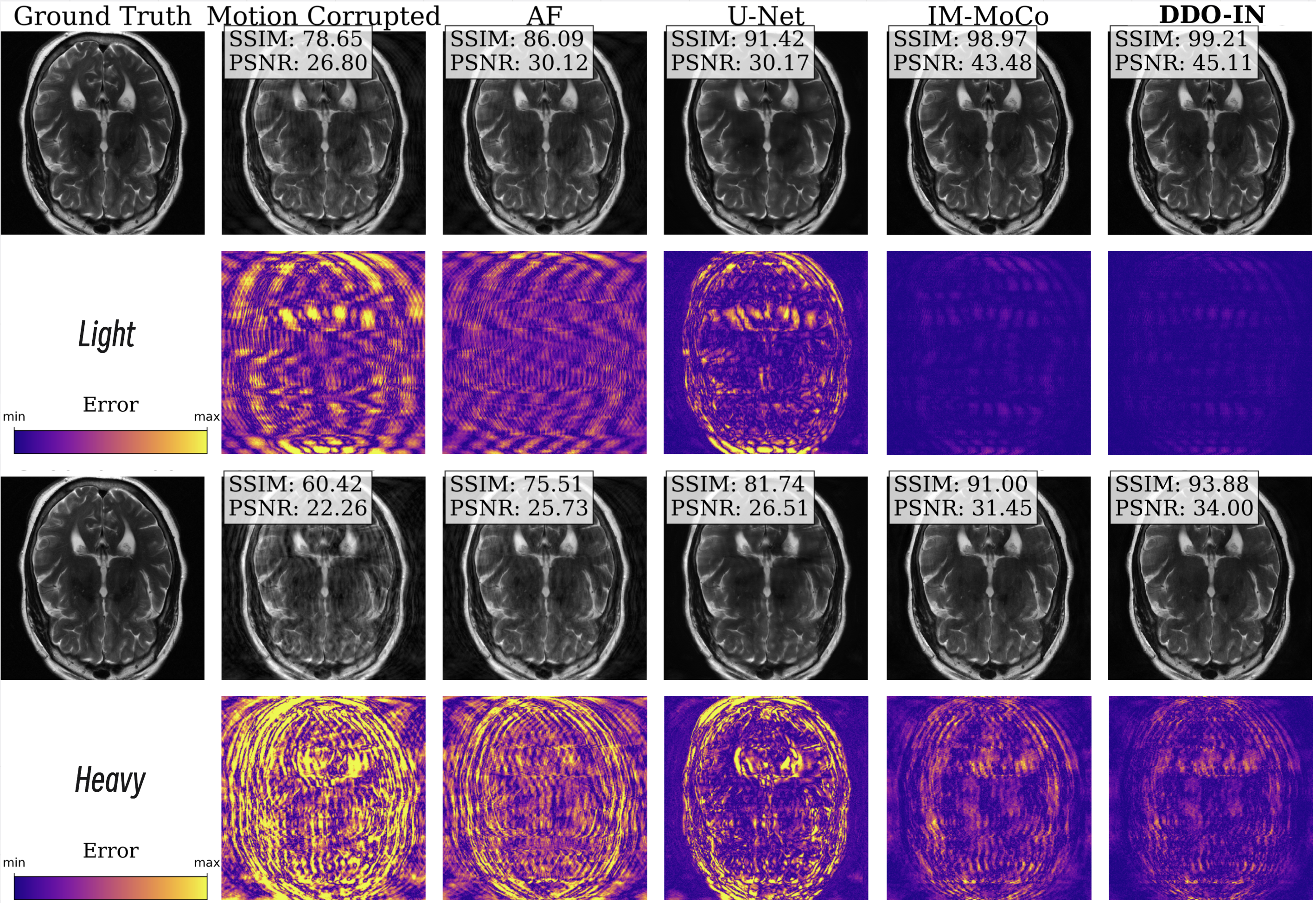}
	\caption{The illustrated data demonstrates the median results of motion-corrected images from our DDO-IN pipeline, alongside motion-corrupted, ground truth, and comparison methods. The first and third rows present the light and heavy correction outcomes, respectively. The second and fourth rows illustrate the residual error images.}
	\label{fig2}
\end{figure}
\subsection{Results}

The Fig.~\ref{fig2} presents a comparative analysis of motion correction results, illustrating original, motion-corrupted, and corrected images for both mild and severe motion, accompanied by an array of performance metrics. Detailed contours serve to highlight the effectiveness of artifact removal. Our method is systematically benchmarked against unprocessed data and the IM-MoCo approach, considered the current state-of-the-art in the field. The superior performance of IM-MoCo over previous models establishes it as a robust baseline for showcasing the advantages of our method.

As Table.~\ref{tab1} presenting, quantitative comparisons with both IM-MoCo and Motion Corrupted images demonstrate that our DDO-IN method achieves superior performance across mild and severe motion scenarios. In particular, under mild motion conditions, DDO-IN exhibits substantial improvements in the metrics VIF, HaarPSI, PSNR, and SSIM, corresponding to enhancements of 1.47\%, 1.59\%, 0.68 db, and 0.22\%, respectively, when compared to IM-MoCo. Contrasted with Motion Corrupted images, the improvements are even more significant, with increases of 37.46\% in VIF, 31.57\% in HaarPSI, 11.23 db in PSNR, and 21.52\% in SSIM. Likewise, in severe motion scenarios, DDO-IN consistently surpasses IM-MoCo, demonstrating notable metric improvements. Furthermore, the performance of DDO-IN consistently surpasses that of Motion Corrupted images, effectively illustrating its robustness and efficacy under challenging motion conditions. Ultimately, DDO-IN outperforms IM-MoCo across all evaluated metrics, thereby establishing its superior effectiveness in motion correction.

\section{Discussion \& Outlook}
Our experiments highlights the critical role of dynamic loss control through weight adjustments, demonstrating its impact on the focus of the model across different image regions. Notably, linear weight decay proved superior, due to its ability to minimize training oscillations and enhance stability. This study explores integrating frequency and pixel domain information for enhanced model learning. The initial approach, involving immediate summation of domains and subsequent calculation of joint loss, proved inefficient. We shifted to a method where losses for each domain were computed separately before dynamically aggregating them, indicating possible information conflict when applying direct summation. Efficiently leveraging both domains remains a key research direction. While improvements were observed, optimizing this integration remains a key area for future research. The inherent scalability and flexibility of implicit neural representations were also evident, particularly in handling diverse artifacts and their potential for 3D medical image processing.
In the pursuit of future advancements, we intend to investigate sophisticated methodologies for the synergistic integration of frequency and pixel domain information, possibly through the implementation of adaptive fusion strategies. Moreover, further examination of the optimization of weight decay schedules, potentially informed by data-driven approaches, is deemed necessary. The potential of INR in the realm of medical imaging, particularly in three-dimensional applications, necessitates a comprehensive exploration to broaden their applicability across diverse modalities and tasks. This endeavor must concurrently address the computational demands associated with high-resolution imaging.

\section{Conclusion}
This research introduces an innovative DDO-IN pipeline designed to direct INRs in the mitigation of motion artifacts while preserving intricate texture details. The study notably implements alternating masks within the frequency domain and incorporates dynamic weights alongside the pixel domain. Comprehensive experiments conducted on the \emph{NYU fastMRI} dataset validate the exceptional performance of the proposed method. 
\bibliographystyle{splncs04}
\bibliography{mybibliography}
%




\end{document}